\documentclass[sigconf,anonymous=false,twocolumn]{acmart}
\usepackage{diagbox}
\usepackage{subfigure}
\usepackage{caption}
\usepackage{abstract}

\AtBeginDocument{%
  \providecommand\BibTeX{{%
    \normalfont B\kern-0.5em{\scshape i\kern-0.25em b}\kern-0.8em\TeX}}}
    
\makeatletter
\newcommand\settitlebf{%
\def\maketitle{%
    \par\textbf{\@title}%
    \par\@author%
    \par}
}

\author{%
Ye Tang, Xuesong Yang, 
Xinrui Liu,Xiwei Zhao,
Zhangang Lin,Changping Peng
\\[1ex]
 JD Group,Beijing,China\\[0.5ex]
 \{tangye8,yangxuesong1,liuxinrui10,zhaoxiwei,linzhangang,pengchangping\}@jd.com \\[0.5ex]
}

\begin{document}

\title{NDGGNET-A Node Independent Gate based Graph Neural Networks}
\maketitle

\begin{abstract}
Graph Neural Networks (GNNs) is an architecture for structural data, and has been adopted in a mass of tasks and achieved fabulous results, such as link prediction, node classification, graph classification and so on. Generally, for a certain node in a given graph, a traditional GNN layer can be regarded as an aggregation from one-hop neighbors, thus a set of stacked layers are able to fetch and update node status within multi-hops. For nodes with sparse connectivity,  it is difficult to obtain enough information through a single GNN layer as not only there are only few nodes directly connected to them but also can not propagate the high-order neighbor information. However, as the number of layer increases, the GNN model is prone to over-smooth for nodes with the dense connectivity, which resulting in the decrease of accuracy. To tackle this issue, in this thesis, we define a novel framework that allows the normal GNN model to accommodate more layers. Specifically, a node-degree based gate is employed to adjust weight of layers dynamically, that try to enhance the information aggregation ability and reduce the probability of over-smoothing. Experimental results show that our proposed model can effectively increase the model depth and perform well on several datasets.
\end{abstract}


\section{Introduction}
Graph neural network (GNN) is a powerful model that could transmit information along with connections within a given graph. Thus, for each node in a graph, the network collects information from neighbors and deduces its state after several stacked layers. On this mechanism, three famous fantastic improvements are proposed. GCN focus on aggregation and built a fundamental pipeline to combine node-wise state and neighbor-wise state\cite{GCN}. GAT considered that one node should have various dependency levels with neighbors, then infer a set of scores by taking advantages of attention module for nodes updating\cite{GAT}. GraphSage borrowed the idea of random walk\cite{deepwalk} and unveil neighbor sampling instructively, that make GNN model suitable for inductive learning\cite{GraphSAGE}. Inspiring with these three ideas, tremendous related works have been published and adopted to tackle downstream task of all kinds of domains, including computer vision\cite{cvgraph}, nature language process\cite{nlpgraph}, social network analysing\cite{sograph}, medical research\cite{migraph}, recommendation system\cite{regraph} and so on. However, these models usually illuminate heterogeneous capability on graphs with various average number of edges\cite{graphsaint}. 

Generally, in most graph based tasks, it is should be fully presented that each node has limited edge info, in meanwhile, the information of high connectivity node could help the lower one by the correlation between them. In graph neural network, the aggregation of high-order neighboring information in a graph is the key effective way to implement the between-node high correlation topology. When we deploy a deeper graph layer for the low connectivity node, in meanwhile, the active node which has abundant edges between nodes within lower order graph domain will be easily over-fitted as the most common structure of graph neural network is stacking-layer. Take Graph-Sage for instance,  to overcome the probably over-fitting issue, for most state-of-the-art GNN implementation, the layer depth is generally not bigger than three. At the same time, the 3 layer depth is probably far away from the best performance because the imbalance connectivity distribution of different nodes is not take into consideration. However, the deeper GNN framework has been figured out by the inspiration of Res-Net,but the complexity of structure ignores the efficiency of forward propagation when it deepen the depth of the layer. How to deploy the dynamic multi layers of graph neural network by the different characteristic of nodes in a simple and efficient way is necessary.  

On the other hand, a mass of thesis suggest that a GNN model would make all nodes tend to steady state eventually, and such so-called final state is detrimental to downstream tasks as nodes in graph may share similar vectorized expressions and make it indistinguishable\cite{gresnet,NDLS,TowardsDeep}. Therefore, some studies have shown that, residual like architecture can slow the rate of convergence, that make it possible to build a network with numerous layers\cite{GCNII,GCN,gresnet}. And some researchers lay emphasis on implicit relationship between features and graph structures, that spawn attention-wise methods\cite{hopattention,graphomer}. Parameter reduction is also an effective idea for over smooth dilution, and the ensuing framework are able to obtain modest computational complexity by wiping off some weak related parameters or make them shareable\cite{train1000,SGC}.

In this thesis, we first analyze the factors affecting the increasing of network layers,then propose a node independent gate based graph neural network (NDGGNET) for the different connectivity topology node within a certain graph. The model should design an attention based gate structure which can establish the natural relation between over smooth issue and layer depth in propagation and achieve outperformed numerical results on several open datasets. The main contributions is summarized as follow:
\begin{itemize}
    \item we analyse the factors that may related with over smoothing issue;
    \item a node independent gate is introduced to uniformize the operation process.
\end{itemize}

\section{Preliminaries}\label{sec:2}
\subsection{Graph Neural Network}
Give a graph $G(v, e, X)$, where $v$ is node set, $e$ is edge set, and $X$ is initial feature of nodes, graph convolution operator can be illuminated by two steps, i.e. propagating and mapping. Seeing in Eq.1, GNN aggregate intensities of nodes in kth layer, $H^k$, by following the normalized adjacency matrix $\hat{A}$, then extend these features with weight matrix of this layer, $W^k$, and a non-linear activation function $\sigma(\bullet)$
\begin{equation}
\begin{aligned}
&H^{k} = \sigma(\hat{A}H^{k-1}W^k)\\
&H^0 = X
\end{aligned}
\label{eq-gnn}
\end{equation}
\subsection{Over Smoothing}\label{sec:ov}
Following equation in Eq.1, a graph neural network can transmit message among nodes continuously, and update feature of each node according with its neighbor nodes, thus makes every nodes are able to sense information in the graph. However, suggesting by massive of existing papers \cite{GCNII,NDLS}, such framework with sufficient number of layers will force each node to achieve stable and reduce diversity. This phenomenon is called over smoothing.

To simplify the analysis, we assume activation function in Eq. \ref{eq-gnn} as RELU\cite{RELU}, so it can be ignored if input feature values are all non negative, then the calculation formula of each node, $v$, in GNN can be written as expansion in Eq.\ref{eq:extend}, where $W^i$ is weight matrix in i-th layer, $W^* = \prod_{i=0}^n W^i$, $\mathcal{N}(v)$ is set nodes of graph, $G$, $\tilde{A} = A + I$ is normalized adjacent matrix and $\tilde{D}$ is degree matrix with self loop.

 \begin{equation}
 \begin{aligned}
 \lim_{k=0}^{\infty} H^k &= \hat{A}^kX^0W^*\\
 &= (\tilde{D}^{-\frac{1}{2}}\tilde{A} \tilde{D}^{-\frac{1}{2}})^kX^0W^*\\
 &=\frac{\sum_{u\in \mathcal{N}(G)}\sqrt{(d_v+1)(d_u+1)}}{2m+n}x_{v}^0W^*
\end{aligned}
\label{eq:extend}
\end{equation}

Based on this assumption, the final state of one node only related with its own degree, input features and transformation matrixes, that means over smoothing issue could be mitigated if these factors are well processed.

\subsection{Influence of Smoothing}\label{sec:inf}
The above analysis shows that the node features converge with infinity layers. In other words, the distance between nodes becomes very small to result in weak feature classification. In Eq.\ref{eq:MDCN}, we define a mean distance of connected nodes (MDCN) to show the variation of distance, where $x_i$ is feature vector of  $i_{th}$ node and $\|\bullet\|_2^2$ is l2-norm distance. 

\begin{equation}
	MDCN = \frac{1}{n}\sum_{i=1}^N \sum_{j\in \mathcal{N}(i)}\|x_i-x_j\|_2^2
	\label{eq:MDCN}
\end{equation}

\begin{figure}[htbp]
  \centering
  \includegraphics[width=\linewidth]{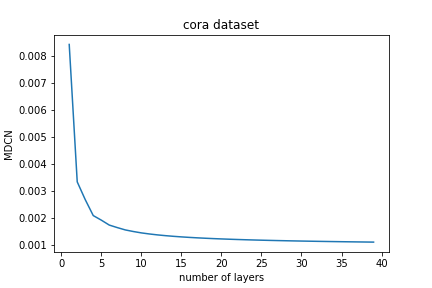}
  \caption{MDCN with defferent layers in cora dataset.}
  \label{fig:influence}
\end{figure}

In Figure \ref{fig:influence} shows for the MDCN of different number layer in cora dataset\cite{Planetoid}, the distance among connected nodes drops dramatically within few layers, and that is a hard evidence to decipher strong correlation between number of layers and over smoothing issue.

\section{Methodology}
In this section, we first give a description of degree based gate and dynamic degree based gate. The model is shown in figure \ref{fig:model}.

\subsection{Intuition\label{sec:inf}}
In order to reveal the relationship between expression decay and node degree, we will conduct the stationary state for each node-i in graph, we now hypothesis the stationary state for any node reaching is K(i,$\epsilon$)($\epsilon$ is a vary small constant express the node will be effected quite limited by other nodes with the increasing iteration), in Eq.4, we found that the K is vary by the different nodes in graph and related to the connectivity of the node itself.

\begin{equation}
\begin{aligned}
K(i,\epsilon) &<= \log_{\lambda_2}(\epsilon\sqrt{\frac{\hat{d_i}}{2m+n}})\\
\end{aligned}
\label{eq4}
\end{equation}

where $\lambda_2$ is the second largest eigenvalue of $\hat{A}$, $\hat{d_i}$ denotes the degree of node $v_i$, and m denote the number of edges and n denotes the number of nodes separately.
It is obviously that the lower connectivity node i pretends to be company with the smaller K(i,$\epsilon$) and the higher node j company with a bigger K(j,$\epsilon$)

\subsection{Fundamental Framework}
\begin{figure}[htbp]
  \centering
  \includegraphics[width=\linewidth]{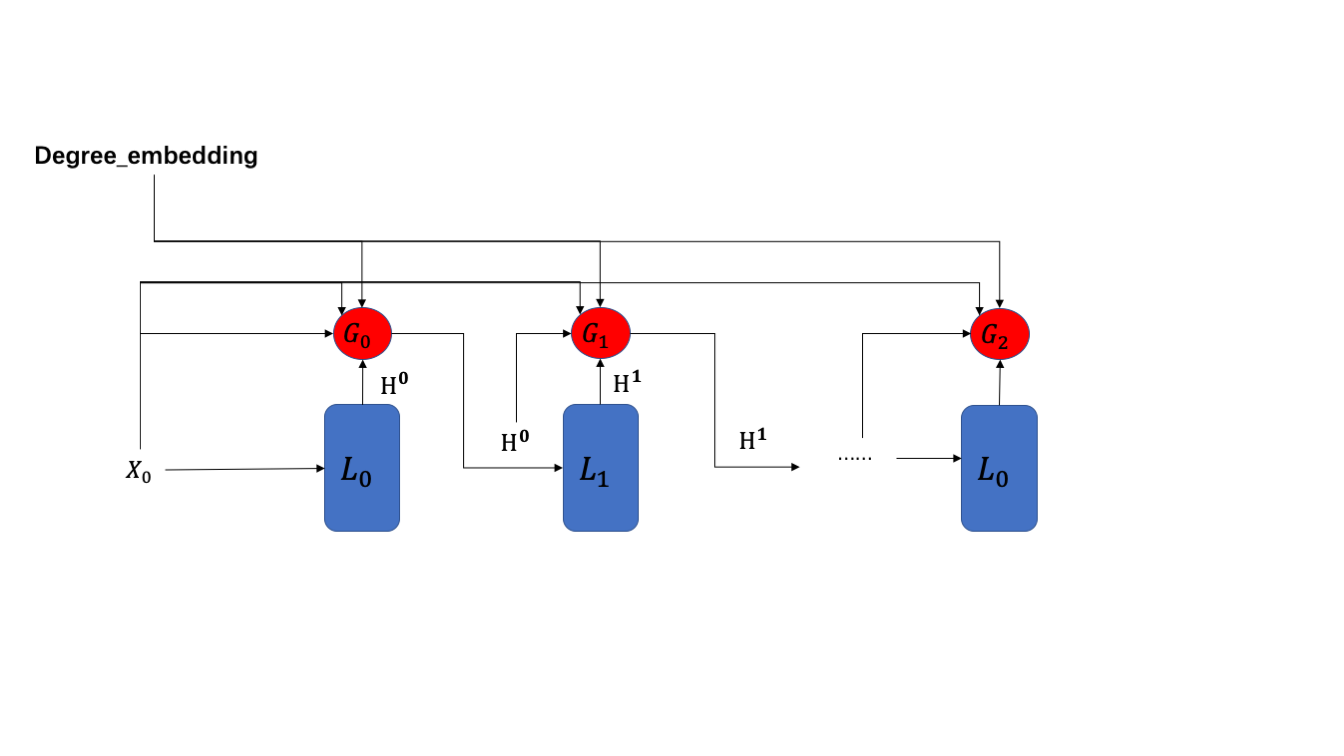}
  \caption{flowchart of the proposed model.}
  \label{fig:model}
\end{figure}

Inspiring by solid work in computer vision\cite{resnet}, we adopt a residual wise framework to slow the variation between layers. Specially, seeing in Eq.\ref{eqresidual}, for $k_{yh}$ layer in GNN model, the output feature of previous layer can be regarded as a momentum term, and a weighted summation is deployed to combine the current term, $\tilde{A}X^{k-1}W^k$ and history term, $X^{k-1}$.
\begin{equation}
\begin{aligned}
&H^k = (1-\alpha^k)\sigma(\hat{A}H^{k-1}) + \alpha^k H^{k-1}\\
&H^0 = \sigma(\hat{A}X^{0})
\end{aligned}
\label{eqresidual}
\end{equation}
Figure.\ref{fig:model} shows basic architecture of our model, that we set a node-wise gate, $\alpha^i$, to balance terms in $i_{th}$ layer.

\subsection{Node Independent Gate}
Generally, weights in moving strategy is high related with model performance, and it is sensitive and uneasy to assign an optimal value artificially. As it is mentioned in front pages, final state of GNN is strongly related with degree of vertex and initial features, it controls convergence rate. Thus, we build a attention gate, seeing in Eq.\ref{eq:gate1}, which takes a concatenation of d-dim degree embedding, $E(D)\in R^{n*d}$, initial feature, $X_0$, output of last layer, $H_{k-1}$, and output of this layer, $H_k$, then a value between 0 and 1 is derived by making use of MLP mapping with softmax, $g(\bullet)$. that balances the residual link between input and output of one GNN layer.

\begin{equation}
\alpha^k = g(E(D)\|X_0\|H_k\|H_{k-1})
\label{eq:gate1}
\end{equation}

\begin{figure}[htbp]
  \centering
  \includegraphics[width=\linewidth]{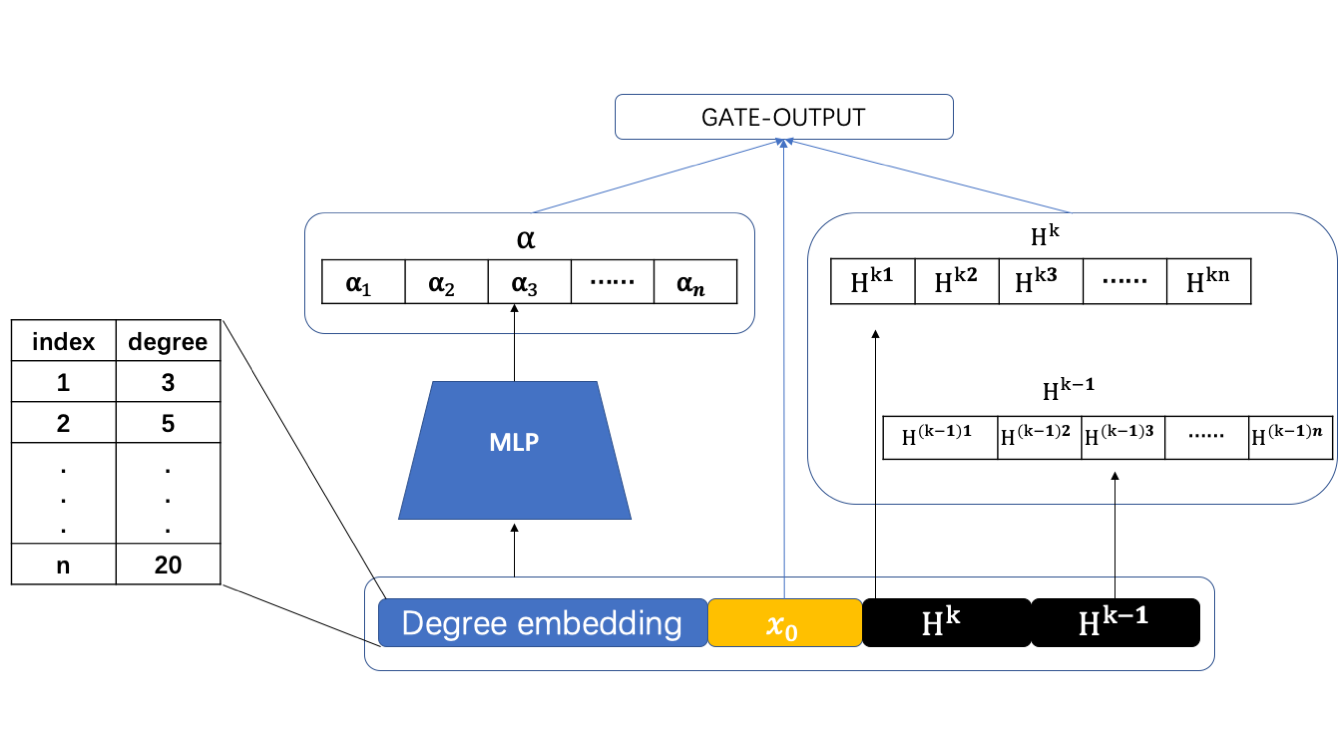}
  \caption{framework of node independent gate.}
  \label{fig:gate}
\end{figure}

Additionally, if Figure.\ref{fig:gate}, we treat elements in gate value of one node separately and independently , that means the output of gate function should be a vector, $\alpha^k = \{\alpha_1^k, \alpha_2^k,..., \alpha_i^k,...,\alpha_c^k\}$, where $c$ is dim size of node feature. Details can be found in section \ref{sec:experimemts}.

\section{Experiments}\label{sec:experimemts}
In this section, we will demonstrate the results of our proposed architecture in contrast with popular GNN methods on open datasets.
\subsection{Implementation Details}
In experiment, NDGGNET based models are well trained by Adam optimizer\cite{Adam} with exponential learning rate decay(LR decay), and other parameters, including weight decay, drop rate, max expoch, are depicted in table.\ref{tab:trainingP}. And for clarity of definition, notation NDGGNET in all tables means our standard proposed method and NDGGNET(*) is a degraded version that the attention scores of both terms in Eq.\ref{eqresidual} is set with 1.

\begin{table}[htbp]
    \centering
    \begin{tabular}{l|c|c|c}
        \hline
        para name&weight decay&dropout rate&max epochs\\
        value&5e-4&0.5&500\\
        \hline
        para name&init LR&LR decay step&LR decay rate\\
        value&1e-2&10&0.95\\
        \hline
    \end{tabular}
    \caption{training parameters setting}
    \label{tab:trainingP}
\end{table}

Moveover, we evaluated proposed method and its variants in contrast with other wide applied GNN methods, i.e. GCN and its decoupled version\cite{GCN}, GAT\cite{GAT}, SGC\cite{SGC}, SIGN\cite{SIGN}, APPNP\cite{APPNP} and JK-NET\cite{JKNET} on three datasets, i.e. cora, citeseer and pubmed\cite{Planetoid}, and accuracy. To be fair, accuracy is employed to measure the effects respectively, details of datasets are shown in table.\ref{tab:datasets}

\begin{table}[htbp]
    \centering
    \begin{tabular}{l|c|c|c|c}
        \hline
        dataset&nodes&edges&features&classes\\
        \hline
        \hline
        cora&2708&5429&1433&7\\
        citeseer&3327&4732&3707&6\\
        pubmed&19717&44338&500&3\\
        \hline
    \end{tabular}
    \caption{Details of open datasets.}
    \label{tab:datasets}
\end{table}
\subsection{Numerical Results}
In table.\ref{tab:Tresults}, the accuracy of our method is significantly improved in compare with both coupled methods and decoupled methods. That proves the effectiveness of the functionality of node independent gating function.

\begin{table}[htbp]
    \centering
    \begin{tabular}{l|c|c|c}
        \hline
        \diagbox{method}{dataset}&cora&citeseer&pubmed\\
        \hline
        GCN&81.5&73.4&79.0\\
        GCN-decoupled&83.0&70.8&78.8\\
        GAT&83.0&72.5&78.8\\
        APPNP&83.3&71.8&\textbf{80.1}\\
        SGC&81.0&71.3&78.9\\
        SIGN&82.1&72.4&79.5\\
        JK-Net&81.8&70.7&78.8\\
        NDGGNET&\textbf{84.3}&\textbf{73.8}&79.0\\
        NDGGNET(*)&81.3&69.0&78.1\\
        \hline
    \end{tabular}
    \caption{Experimental results of graph learning. Take accuracy as metrics.}
    \label{tab:Tresults}
\end{table}

\begin{figure}[htbp]
  \centering

    \subfigure{\includegraphics[width=0.8\columnwidth]{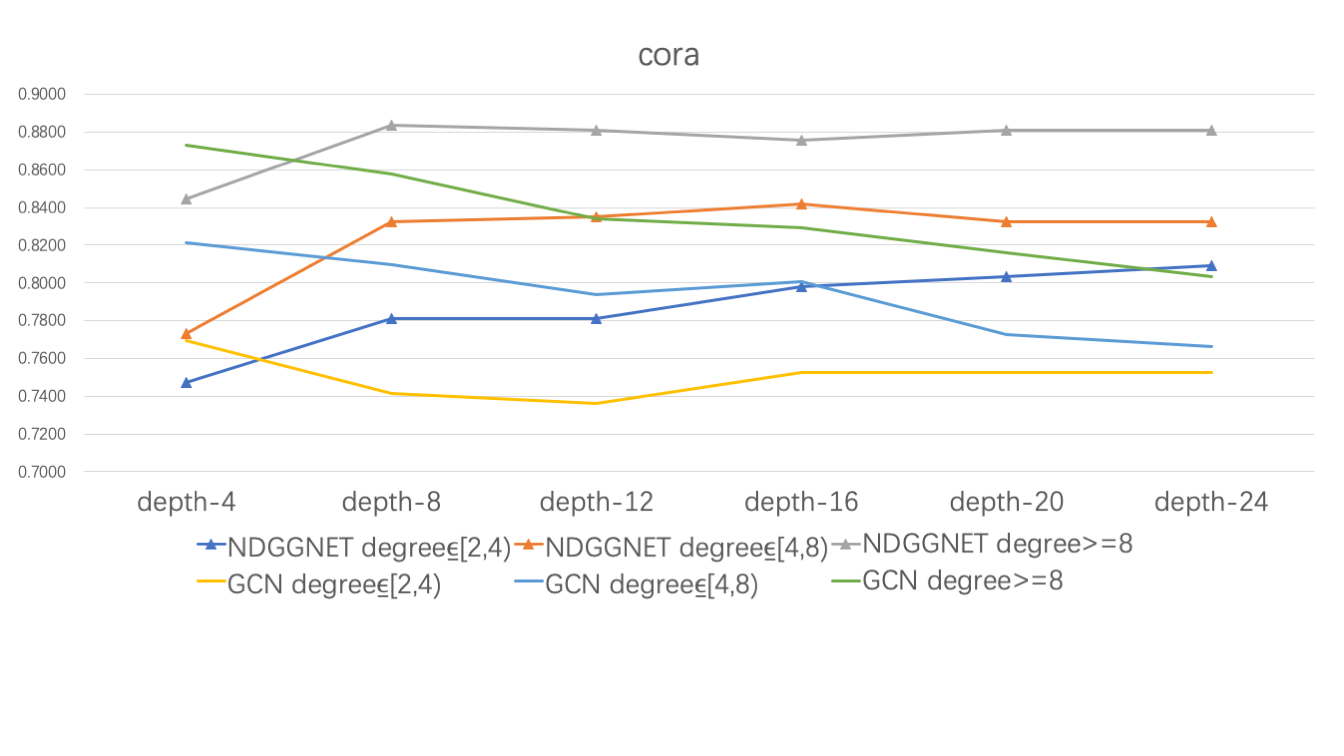}} \\

    \subfigure{\includegraphics[width=0.8\columnwidth]{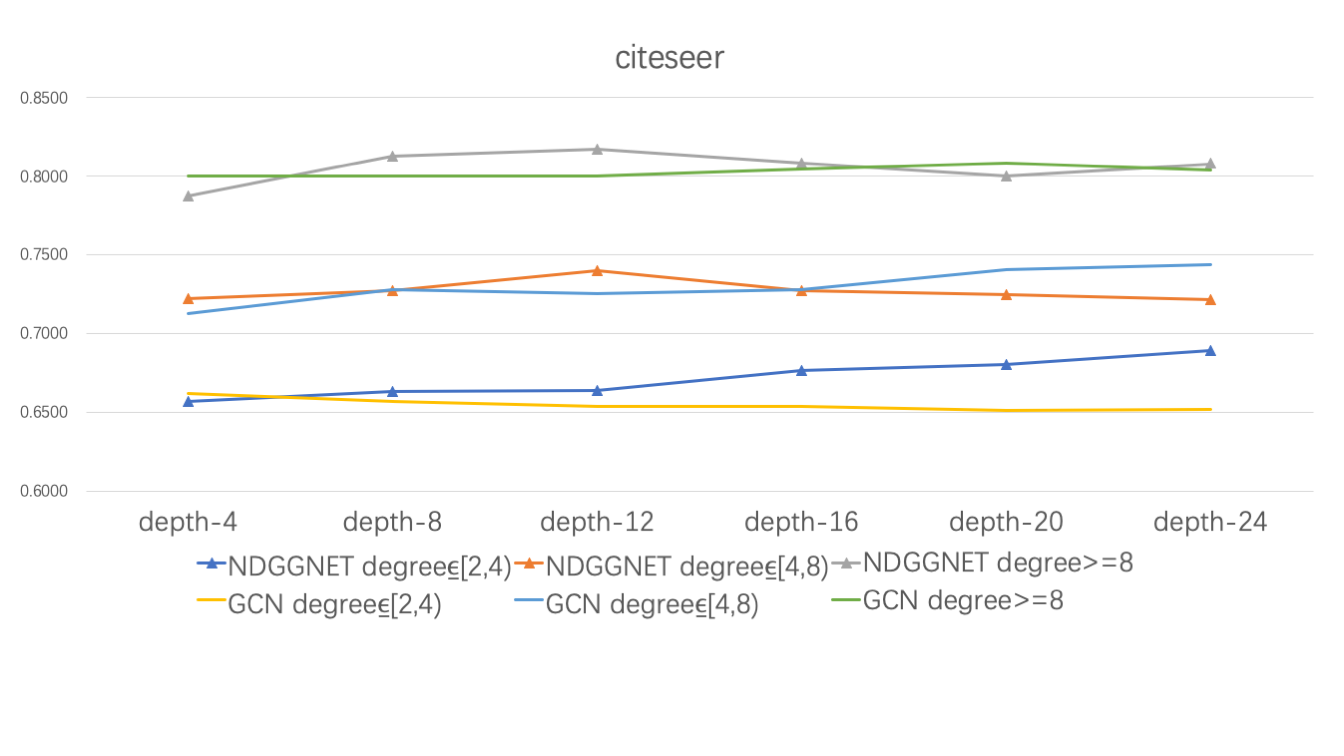}} \\

    \subfigure{\includegraphics[width=0.8\columnwidth]{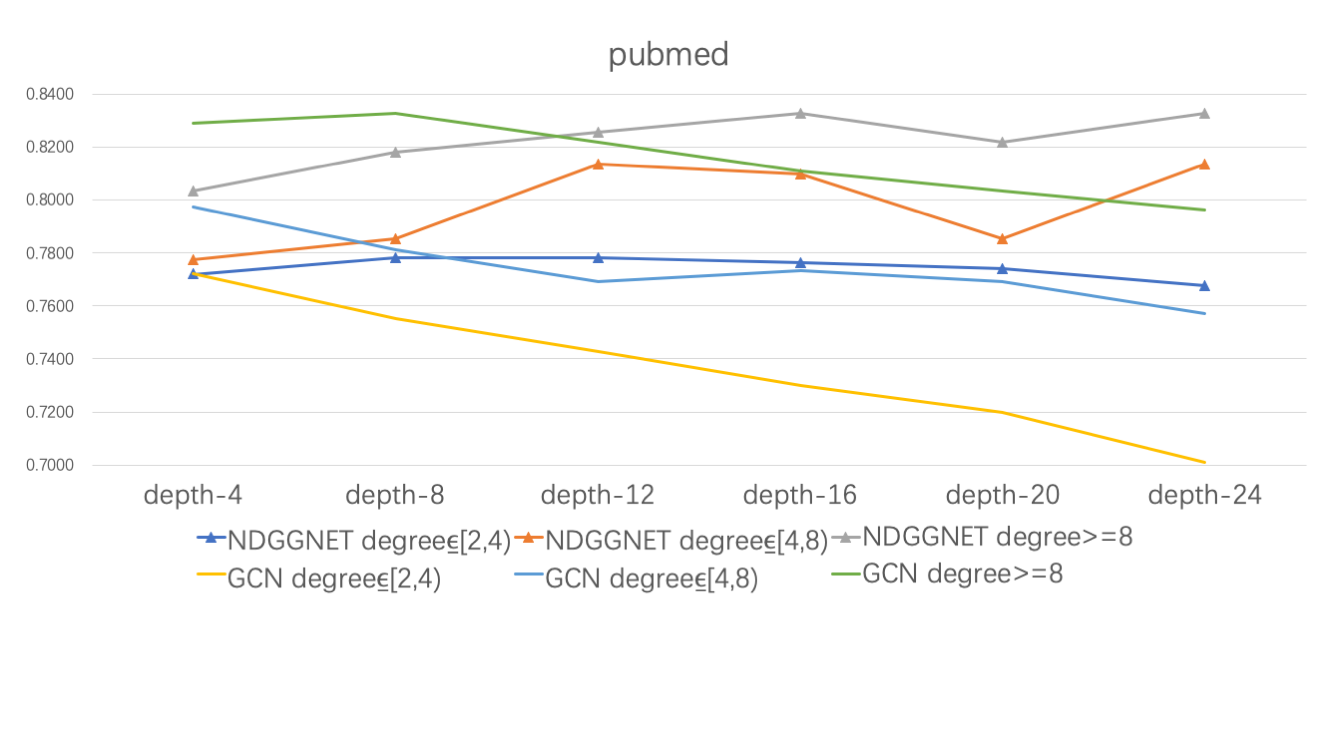}} \\
    
  \caption{Accuracy of traditional GCN model and NDGGNET with different node degree and number of layers.}
  \label{fig:depth}
\end{figure}

Further, we adopt traditional GCN model and NDGNNET, and divide each dataset into three barrels according node degree, i.e. $deg\in[2,4),deg\in[4,8),deg\in[8,\infty)$. Therefore, in Figure.\ref{fig:depth}, the relationships between number of layers, connectivity of nodes and test accuracy are exhibited clearly, each subfigure contains six plots and each plot stands for one GNN method and one barrel. Accuracy of GCN network will drop apace after few layers, but in NDGGNET the accuracy could increase along with layer's number. Specifically, for nodes with sparse connected edges, the NDGGNET model could help them to achieve higher accuracy value by adding more layers, and for nodes with dense connected edges, the accuracy could maintain high values. 

\section{Conclusion}
In this thesis, we first analyze the nature of over smooth, show the cause of the issue, and deduce the relevant factors, then propose a node independent gate function to control the rate of transmission of the GNN network, that make it possible to dilute over smooth and expend the number of layers. The experimental results illuminate the designed gate could regulate the increment in updating of nodes adaptively. However, our method does not excavate the relationships between terms in gate attention deeply, thus leaves a lot of room for improvement, we will give a more profound research in 
.


\begin{thebibliography}{10}

\bibitem{RELU}
Jason Brownlee.
\newblock A gentle introduction to the rectified linear unit (relu).
\newblock {\em Machine learning mastery}, 6, 2019.

\bibitem{GCNII}
Ming Chen, Zhewei Wei, Zengfeng Huang, Bolin Ding, and Yaliang Li.
\newblock Simple and deep graph convolutional networks.
\newblock {\em CoRR}, abs/2007.02133, 2020.

\bibitem{sograph}
Lina Gong, Gopi Krishnan~Krishnan Rajbahadur, Ahmed~E. Hassan, and S.~Jiang.
\newblock Revisiting the impact of dependency network metrics on software
  defect prediction.
\newblock {\em IEEE Transactions on Software Engineering}, page 1–1, 2021.

\bibitem{nlpgraph}
Abdullah Hamid, Nasrullah Sheikh, Naina Said, Kashif Ahmad, Asma Gul, Laiq
  Hassan, and Ala~I. Al{-}Fuqaha.
\newblock Fake news detection in social media using graph neural networks and
  {NLP} techniques: {A} {COVID-19} use-case.
\newblock {\em CoRR}, abs/2012.07517, 2020.

\bibitem{GraphSAGE}
William~L. Hamilton, Rex Ying, and Jure Leskovec.
\newblock Inductive representation learning on large graphs.
\newblock {\em CoRR}, abs/1706.02216, 2017.

\bibitem{cvgraph}
Guangxing Han, Yicheng He, Shiyuan Huang, Jiawei Ma, and Shih{-}Fu Chang.
\newblock Query adaptive few-shot object detection with heterogeneous graph
  convolutional networks.
\newblock {\em CoRR}, abs/2112.09791, 2021.

\bibitem{resnet}
Kaiming He, Xiangyu Zhang, Shaoqing Ren, and Jian Sun.
\newblock Deep residual learning for image recognition.
\newblock {\em CoRR}, abs/1512.03385, 2015.

\bibitem{migraph}
Yining Huang, Meilian Chen, and Keke Tang.
\newblock Training like playing: {A} reinforcement learning and knowledge
  graph-based framework for building automatic consultation system in medical
  field.
\newblock {\em CoRR}, abs/2106.07502, 2021.

\bibitem{Adam}
Diederik~P Kingma and Jimmy Ba.
\newblock Adam: A method for stochastic optimization.
\newblock {\em arXiv preprint arXiv:1412.6980}, 2014.

\bibitem{GCN}
Thomas~N. Kipf and Max Welling.
\newblock Semi-supervised classification with graph convolutional networks.
\newblock {\em CoRR}, abs/1609.02907, 2016.

\bibitem{APPNP}
Johannes Klicpera, Aleksandar Bojchevski, and Stephan G{\"{u}}nnemann.
\newblock Personalized embedding propagation: Combining neural networks on
  graphs with personalized pagerank.
\newblock {\em CoRR}, abs/1810.05997, 2018.

\bibitem{regraph}
Ansong Li, Zhiyong Cheng, Fan Liu, Zan Gao, Weili Guan, and Yuxin Peng.
\newblock Disentangled graph neural networks for session-based recommendation.
\newblock {\em CoRR}, abs/2201.03482, 2022.

\bibitem{train1000}
Guohao Li, Matthias M{\"{u}}ller, Bernard Ghanem, and Vladlen Koltun.
\newblock Training graph neural networks with 1000 layers.
\newblock {\em CoRR}, abs/2106.07476, 2021.

\bibitem{TowardsDeep}
Meng Liu, Hongyang Gao, and Shuiwang Ji.
\newblock Towards deeper graph neural networks.
\newblock {\em CoRR}, abs/2007.09296, 2020.

\bibitem{deepwalk}
Bryan Perozzi, Rami Al{-}Rfou, and Steven Skiena.
\newblock Deepwalk: Online learning of social representations.
\newblock {\em CoRR}, abs/1403.6652, 2014.

\bibitem{SIGN}
Emanuele Rossi, Fabrizio Frasca, Ben Chamberlain, Davide Eynard, Michael~M.
  Bronstein, and Federico Monti.
\newblock {SIGN:} scalable inception graph neural networks.
\newblock {\em CoRR}, abs/2004.11198, 2020.

\bibitem{hopattention}
Chuxiong Sun and Guoshi Wu.
\newblock Adaptive graph diffusion networks with hop-wise attention.
\newblock {\em CoRR}, abs/2012.15024, 2020.

\bibitem{GAT}
Petar Veličković, Guillem Cucurull, Arantxa Casanova, Adriana Romero, Pietro
  Liò, and Yoshua Bengio.
\newblock Graph attention networks.
\newblock 2018.

\bibitem{SGC}
Felix Wu, Tianyi Zhang, Amauri H.~Souza Jr., Christopher Fifty, Tao Yu, and
  Kilian~Q. Weinberger.
\newblock Simplifying graph convolutional networks.
\newblock {\em CoRR}, abs/1902.07153, 2019.

\bibitem{JKNET}
Keyulu Xu, Chengtao Li, Yonglong Tian, Tomohiro Sonobe, Ken{-}ichi
  Kawarabayashi, and Stefanie Jegelka.
\newblock Representation learning on graphs with jumping knowledge networks.
\newblock {\em CoRR}, abs/1806.03536, 2018.

\bibitem{Planetoid}
Zhilin Yang, William~W. Cohen, and Ruslan Salakhutdinov.
\newblock Revisiting semi-supervised learning with graph embeddings.
\newblock {\em CoRR}, abs/1603.08861, 2016.

\bibitem{graphomer}
Chengxuan Ying, Tianle Cai, Shengjie Luo, Shuxin Zheng, Guolin Ke, Di~He,
  Yanming Shen, and Tie{-}Yan Liu.
\newblock Do transformers really perform bad for graph representation?
\newblock {\em CoRR}, abs/2106.05234, 2021.

\bibitem{graphsaint}
Hanqing Zeng, Hongkuan Zhou, Ajitesh Srivastava, Rajgopal Kannan, and Viktor~K.
  Prasanna.
\newblock Graphsaint: Graph sampling based inductive learning method.
\newblock {\em CoRR}, abs/1907.04931, 2019.

\bibitem{gresnet}
Jiawei Zhang and Lin Meng.
\newblock Gresnet: Graph residual network for reviving deep gnns from suspended
  animation.
\newblock {\em CoRR}, abs/1909.05729, 2019.

\bibitem{NDLS}
Wentao Zhang, Mingyu Yang, Zeang Sheng, Yang Li, Wen Ouyang, Yangyu Tao, Zhi
  Yang, and Bin Cui.
\newblock Node dependent local smoothing for scalable graph learning.
\newblock {\em CoRR}, abs/2110.14377, 2021.

\end{thebibliography}
\bibliographystyle{ACM-Reference-Format}
\end{document}